\documentclass[runningheads]{llncs}
\usepackage{graphicx}

\usepackage{tikz}
\usepackage{comment}
\usepackage{amsmath,amssymb} 
\usepackage{color}

\usepackage{hhline}
\usepackage{multirow}
\usepackage{makecell}
\usepackage{graphicx}
\usepackage{booktabs}

\usepackage{wrapfig}

\makeatletter
\newcommand\footnoteref[1]{\protected@xdef\@thefnmark{\ref{#1}}\@footnotemark}
\makeatother

\usepackage{xcolor,colortbl}
\newcommand{\first}[1]{\cellcolor{yellow!75}\textbf{#1}}
\newcommand{\second}[1]{\cellcolor{yellow!25}{#1}}

\newcommand{\JS}[1]{\textcolor{blue}{#1}}

\usepackage{xspace}

\makeatletter
\newcommand{\plusplus}{\texttt{+}\texttt{+}\xspace}
\makeatother

\usepackage{xcolor,pifont}
\newcommand*\colourcheck[1]{%
  \expandafter\newcommand\csname #1check\endcsname{\textcolor{#1}{\ding{52}}}%
}
\colourcheck{green}

\newcommand*\colourxmark[1]{%
  \expandafter\newcommand\csname #1xmark\endcsname{\textcolor{#1}{\ding{55}}}%
}
\colourxmark{red}

\usepackage{ wasysym }

\makeatletter
\DeclareRobustCommand\onedot{\futurelet\@let@token\@onedot}
\def\@onedot{\ifx\@let@token.\else.\null\fi\xspace}
\def\eg{\emph{e.g}\onedot} 
\def\ie{\emph{i.e}\onedot}

\def\etal{\emph{et al}\onedot}

\newcommand{\Tref}[1]{Table~\textcolor{red}{\ref{#1}}}
\newcommand{\Eref}[1]{Eq.~(\textcolor{red}{\ref{#1}})}
\newcommand{\Fref}[1]{Fig.~\textcolor{red}{\ref{#1}}}
\newcommand{\Sref}[1]{Sec.~\textcolor{red}{\ref{#1}}}

\makeatletter
\def\blfootnote{\gdef\@thefnmark{}\@footnotetext}
\makeatother

\usepackage[width=122mm,left=12mm,paperwidth=146mm,height=193mm,top=12mm,paperheight=217mm]{geometry}

\begin{document}
\pagestyle{headings}
\mainmatter
\def\ECCVSubNumber{4588}

\title{PointMixer: MLP-Mixer for \\Point Cloud Understanding}



\author{
Jaesung Choe$^{\dagger}$\inst{1} \and 
Chunghyun Park$^{\dagger}$\inst{2} \and \\ 
Francois Rameau\inst{1} \and 
Jaesik Park\inst{2} \and 
In So Kweon\inst{1}
}
\institute{
Korea Advanced Institute of Science and Technology (KAIST), South Korea
\and
Pohang University of Science and Technology (POSTECH), South Korea \\
\url{https://github.com/LifeBeyondExpectations/ECCV22-PointMixer} \\
\email{$\{$jaesung.choe, frameau, iskweon77$\}$@kaist.ac.kr \\ $\{$p0125ch, jaesik.park$\}$@postech.ac.kr}
}
\authorrunning{Choe et al.}

\maketitle

\vspace{-4mm}
\begin{abstract}
MLP-Mixer has newly appeared as a new challenger against the realm of CNNs and Transformer. Despite its simplicity compared to Transformer, the concept of channel-mixing MLPs and token-mixing MLPs achieves noticeable performance in image recognition tasks. Unlike images, point clouds are inherently sparse, unordered and irregular, which limits the direct use of MLP-Mixer for point cloud understanding. To overcome these limitations, we propose \textbf{PointMixer}, a universal point set operator that facilitates information sharing among unstructured 3D point cloud. By simply replacing token-mixing MLPs with Softmax function, PointMixer can ``mix" features within/between point sets. By doing so, PointMixer can be broadly used for intra-set, inter-set, and hierarchical-set mixing. We demonstrate that various channel-wise feature aggregation in numerous point sets is better than self-attention layers or dense token-wise interaction in a view of parameter efficiency and accuracy. Extensive experiments show the competitive or superior performance of PointMixer in semantic segmentation, classification, and reconstruction against Transformer-based methods. \blfootnote{$\dagger$ Both authors have equally contributed to this work.
}

\end{abstract}

\section{Introduction}
\label{sec:Introduction}
3D scanning devices, such as LiDAR or RGB-D sensors, are widely used to capture a scene as 3D point clouds. Unlike images, point clouds are inherently sparse, unordered, and irregular. These properties make standard neural network architectures~\cite{resnet,vgg} hardly applicable. To tackle these challenges, there have been numerous ad hoc solutions, such as sparse convolution networks~\cite{mink,spconv}, graph neural networks~\cite{point-graph-00,point-graph-01,point-graph-02,graph-attention,edgeconv,point-graph-03,choe2021volumetric}, and point convolution networks~\cite{interpconv,kpconv,deep-parametric,xu2021paconv}. Despite their structural differences, these techniques have all been designed to extract meaningful feature representation from point clouds~\cite{point-cloud-analysis}. 
Among existing solutions, Transformer~\cite{point-recon,point-cloud-transformer,transformer,point-transformer,park2022fast} appears to be particularly beneficial to extract features from point clouds. Indeed, the self-attention layer that encompasses the dense token-wise relations is specifically relevant in the context of processing irregular and unordered 3D points. 
%
%

Beyond the well established realm of CNNs and Transformer, MLP-Mixer~\cite{mlp-mixer} proposes a new architecture that exclusively uses MLPs. Based on the pioneering study~\cite{mlp-mixer}, concurrent works~\cite{cyclemlp,hire-mlp,vision-permutator,convmlp,gmlp,res-mlp,spatial-sift-mlp,spatial-sift-mlp-v2} address the locality issue of MLPs~\cite{cyclemlp,repmlpnet,vision-permutator,as-mlp,spatial-sift-mlp,spatial-sift-mlp-v2} and discuss the necessity of self-attention~\cite{gmlp,synthesizer,morphmlp}. 
More recently, Yu~\etal~\cite{metaformer} claim that the general architecture formulation is more important than the specific token-wise interaction strategies, such as self-attention and token-mixing MLPs.
Despite an increasing interest, the MLP-like architectures for point clouds have not yet been fully explored.


\begin{wrapfigure}{r}{0.50\textwidth}
\vspace{-8mm}
\centering
\includegraphics[width=1.00\linewidth]{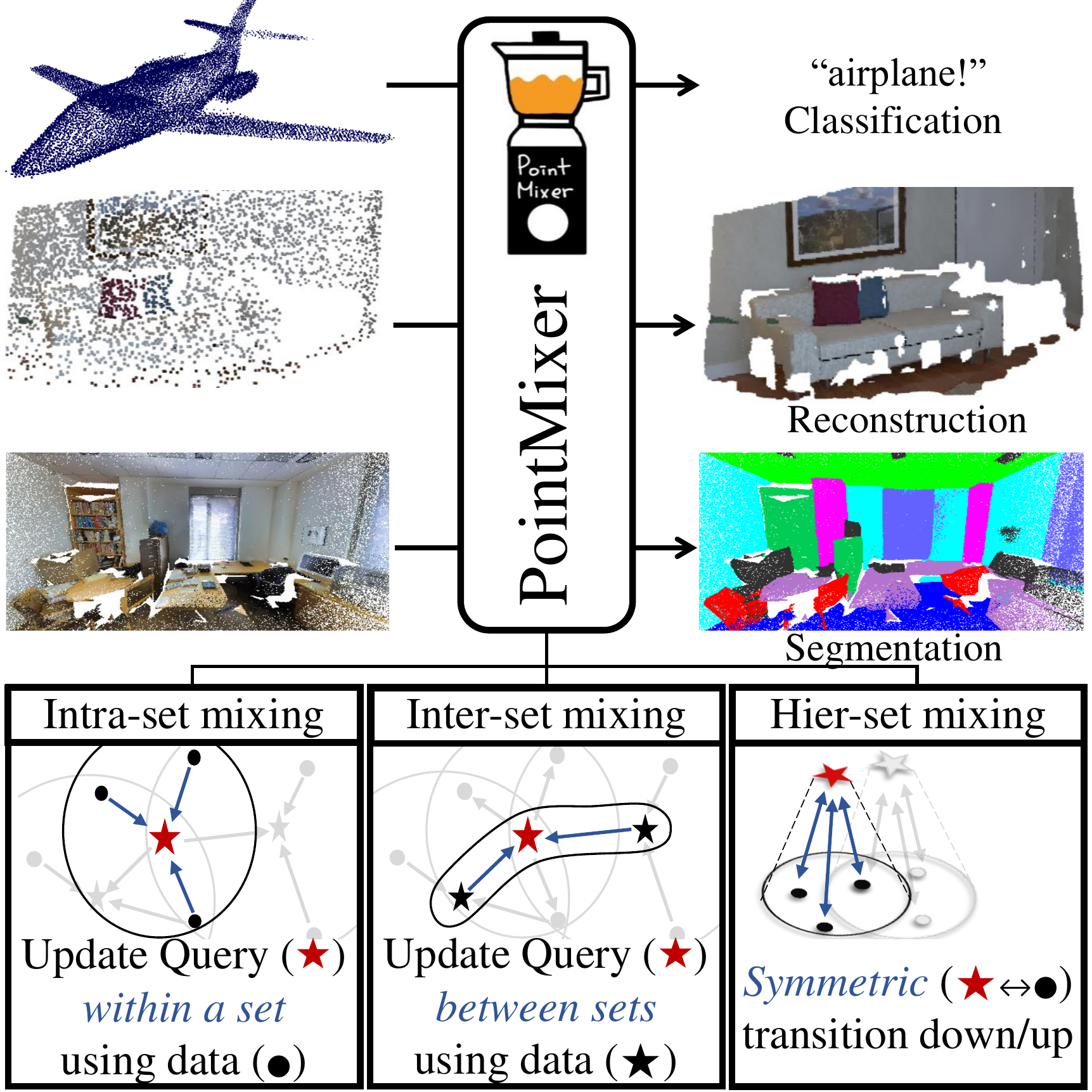}
\vspace{-8mm}
\caption{We present a MLP-like architecture for various point cloud processing, which considers numerous point sets with larger receptive fields to ``mix" information.}
\label{fig:fig_teaser}
\vspace{-6mm}
\end{wrapfigure}
In this paper, we introduce the \emph{PointMixer} that newly extends the philosophy of MLP-like architectures to point cloud analysis. Specifically, we demonstrate the dense token-wise interaction are not essential factors in this context. 
Instead of using token-mixing MLPs, we extend the usage of channel-mixing MLPs into numerous point sets.
As illustrated in~\Fref{fig:fig_teaser} and~\Tref{table:compare}, PointMixer layer shares and mixes point features (1)~within grouped points (intra-set), (2)~between point sets (inter-set), or (3)~points in different hierarchical sets. 
In particular, we newly introduce the concepts of the inter-set mixing and the hierarchical-set mixing, which is clearly different from previous studies~\cite{pointmlp,pointnet++,point-transformer} that only focus on the intra-set mixing. 
To this end, PointMixer layer is a universal point set operator that can propagate point responses into various point sets. 
We claim that various channel MLPs on numerous point sets can outperform self-attention layers or token-mixing MLPs for point clouds.
Moreover, the PointMixer network is a general architecture that has symmetric encoder-decoder blocks fully equipped with PointMixer layers. 

We conduct extensive experiments on various 3D tasks.
These large-scale experiments demonstrate that our method achieves compelling performance among Transformer-based studies~\cite{cloud-transformer,point-transformer}. Our contributions are summarized as below:
%
%
\begin{itemize}
\renewcommand{\labelitemi}{$\bullet$}
    \item PointMixer as a universal point set operator that facilitates mixing of point features through various point sets: intra-set, inter-set, and hierarchical-set. 
    \item Highly accurate and parameter-efficient block design that purely consists of channel-mixing MLPs without token-mixing MLPs.
    \item Symmetric encoder-decoder network to propagate hierarchical point responses through PointMixer layer, instead of trilinear interpolation.
    \item Extensive experiments in various 3D point cloud tasks that highlight the efficacy of PointMixer network against recent Transformer-based studies. 
\end{itemize}


\section{Related Work}
\label{sec:Related work}
\begin{table}[!t]
\centering
\caption{Technical comparisons. Locality represents the local feature aggregation among sampled points. We split the function of the set operator as ``intra"-set mix, ``inter"-set mix, and ``hierarchical"-set mix. Also, we present the symmetric property of encoder-decoder architecture of related work~\cite{pointnet,pointnet++,point-transformer}.
}
\vspace{-3mm}
\resizebox{0.90\linewidth}{!}{
\begin{tabular}{lcccccc}

\Xhline{4\arrayrulewidth} 

\multirow{2}{*}[-0.5ex]{Method} & \multirow{2}{*}[-0.5ex]{~~~Locality~~~} & \multicolumn{3}{c}{~Set operator~} & ~~~~~Symmetric~~~~~ & \multirow{2}{*}[-0.5ex]{~Token-mix~} 
\\ \cline{3-5}

 &  & Intra & ~Inter~ & Hier & pyramid arch & 
\\ \Xhline{2\arrayrulewidth} 
 
PointNet~\cite{pointnet} & \redxmark & \redxmark & \redxmark & \redxmark & \redxmark & Pooling 
\\ 

PointNet\plusplus~\cite{pointnet++} & \greencheck & \greencheck & \redxmark & \redxmark & \redxmark & Pooling 
\\ 


PointTrans~\cite{point-transformer} & \greencheck & \greencheck & \redxmark & \redxmark & \redxmark & Self-attn 
\\ 

PointMLP~\cite{pointmlp} & \greencheck & \greencheck & \redxmark & \redxmark & \redxmark & Affine 
\\ 


\rowcolor{yellow!20} PointMixer (ours) & \greencheck & \greencheck & \greencheck & \greencheck & \greencheck & Softmax 
\\ \Xhline{4\arrayrulewidth} 

\end{tabular}
}
\label{table:compare}
\vspace{-3mm}
\end{table}
In this section, we revisit previous approaches for point cloud understanding, and then briefly introduce Transformers and MLP-like architectures.

\noindent \textbf{Deep learning on point clouds.} \
Point clouds are naturally sparse, unordered, and irregular, which makes it difficult to design a deep neural network for point cloud understanding. To handle such complex data structures, two distinct philosophies have been investigated: voxel-based and point-based methods.

Voxel-based methods~\cite{lee2021putting,mink,3DSemanticSegmentationWithSubmanifoldSparseConvNet,maturana2015voxnet,su2018splatnet,zhou2018voxelnet} first quantize an irregular point cloud into the regular voxel grids, which makes it efficient to search neighbor voxels. 
However, the voxelization process inevitably loses the geometric details of the original point cloud. This issue often leads to infer inaccurate predictions though several recent methods~\cite{liu2019point,tang2020searching,zhang2020deep} try to alleviate the quantization artifacts. Therefore, point-based methods have been actively studied.

PointNet~\cite{pointnet} is a pioneering paper that processes an unordered and irregular points in deep neural architectures. Based on this seminal work, PointNet\plusplus~\cite{pointnet++} presents the ways of involving feature hierarchy as well as points' locality. In details, this paper adopts $k$-Nearest~Neighbor~($k$NN) for local neighborhood sampling and the Farthest Point Sampling algorithm (FPS) for feature hierarchy (\eg, transition downsampling). 
This pyramid encoder-decoder network largely influences on the MLP-based methods~\cite{qian2021assanet,zhao2019pointweb,jiang2019hierarchical,liu2019densepoint,hu2020randla,klokov2017escape,li2018so,huang2018recurrent,xu2020geometry,yang2020cn}, 
%
%
point convolution studies~\cite{komarichev2019cnn,li2018pointcnn,lin2020fpconv,liu2019relation,interpconv,wu2019pointconv,xu2021paconv}, and graph-based networks~\cite{point-graph-00,point-graph-01,edgeconv,graph-attention,point-graph-03}. 
However, as stated in~\Tref{table:compare}, PointNet\plusplus is limited to capture local point responses within the scope of intra-set. Moreover, we found that this vanilla architecture is asymmetric between transition down layers and transition up layers. While downsampling layers adopt pooling with $k$NN and FPS, upsampling layers re-compute $k$NN and trilinear interpolation, which brings asymmetric feature propagation in the encoder-decoder architecture. More recently, Transformer-based study for point cloud processing~\cite{point-transformer} still suffers from the same issue in the pyramid architecture design. 


Our work unifies a local feature aggregation layer, a downsampling layer, and an upsampling layer into \textit{an universal set operator}, named PointMixer layer. This novel layer brings the symmetric and learnable down/upsampling architecture for various 3D perception tasks such as object shape classification~\cite{modelnet40}, semantic segmentation~\cite{armeni_cvpr16} and point cloud reconstruction tasks~\cite{point-recon}.



\noindent \textbf{Transformers and MLP-like architectures.} \
Transformer-based architecture has recently become a game changer in both natural language processing~\cite{devlin2019bert,radford2018improving,synthesizer,transformer} and computer vision~\cite{bello2020lambdanetworks,ViT,han2021transformer,liu2021swin,mao2021voxel,ramachandran2019stand,touvron2021training,point-transformer,yuan2021tokens}.

Vision Transformer (ViT)~\cite{ViT} has opened the applicability of Transformers on visual recognition tasks.
%
%
Because of the quadratic runtime of the self-attention layers
in Transformers, ViT adopts tokenized inputs that divide the image into small region of patches~\cite{trockman2022patches}. The idea of patch embeddings is widely used in the following studies that focus on various issues in ViT~\cite{khan2021transformers,tay2020efficient}: locality~\cite{ramachandran2019stand,vaswani2021scaling,d2021convit,hu2019local} and hierarchy~\cite{wang2021pyramid,xie2021segformer,liu2021swin,chen2021crossvit}.
%
%
With those self-attention layers~\cite{ramachandran2019stand,zhao2020exploring},
Transformer-based point cloud studies~\cite{point-transformer,point-cloud-transformer,cloud-transformer}
demonstrate accurate predictions in both 3D shape classification~\cite{modelnet40} and semantic segmentation~\cite{armeni_cvpr16}. However, a general-purpose layer for both local feature learning and down/upsampling has drawn little attention in handling 3D points.
%
%

%
%
Recently, there exist trials to go beyond the hegemony of CNNs and Transformer by introducing MLP-like architectures. The pioneering paper, MLP-Mixer~\cite{mlp-mixer}, presents a MLP-like network constituted of token-mixing MLPs and channel-mixing MLPs for image classification task. Especially in computer vision, this MLP-like architectures appear to be a new paradigm with their simple formulation and superior performance given large-scale training data. 
%
%
Subsequent papers raise issues and develop potentials in MLP-like architectures: 
(1)~can MLPs handle position-sensitive information or locality~\cite{as-mlp,vision-permutator,spatial-sift-mlp,cyclemlp,repmlpnet}? 
and (2)~does self-attention is truly needed~\cite{synthesizer,gmlp,morphmlp}? Though these issues are still controversial, more recent paper, Metaformer~\cite{metaformer}, addresses the importance of general architecture formulation instead of the specific dense token-wise interaction strategies such as self-attention layers~\cite{transformer,ViT} or token-mixing MLPs~\cite{mlp-mixer}. Simply, by replacing complicated token-mixing operators with the average pooling layer, MetaFormer achieves remarkable performance against the recent MLP-based and Transformer-based studies.
%
%

%
%
Despite their success, modern MLP-like approaches have not yet been fully exploited to point clouds. In contrast to recent Transformer-based point cloud studies~\cite{point-transformer,cloud-transformer,point-cloud-transformer,park2022fast}, our PointMixer network is a general-purpose architecture that symmetrically upsamples/downsamples points' responses and truly exploits the strength of MLPs to operate mixing within/beyond sets of points. By doing so, we successfully conduct various tasks, 3D semantic segmentation, point cloud reconstruction, and object classification tasks~\cite{armeni_cvpr16,point-recon,modelnet40}.


\section{Method}
\label{sec:PointMixer}
In this section, we describe the details of our PointMixer design. For the sake of clarity, we compare the general formulation of MLP-Mixer with representative point-based approaches, such as PointNet\plusplus and Point Transformer (\Sref{subsec:preliminary}). Then, we examine whether MLP-Mixer is of relevance to a point set operator (\Sref{subsec:MLP-Mixer as Point Set Operator}). Finally, we introduce our PointMixer layer (\Sref{subsec:Point Mixer Layer}) that is adopted in our entire network (\Sref{subsec:Network Architecture}).

\begin{figure}[!t]
\centering
\includegraphics[width=0.88\linewidth]{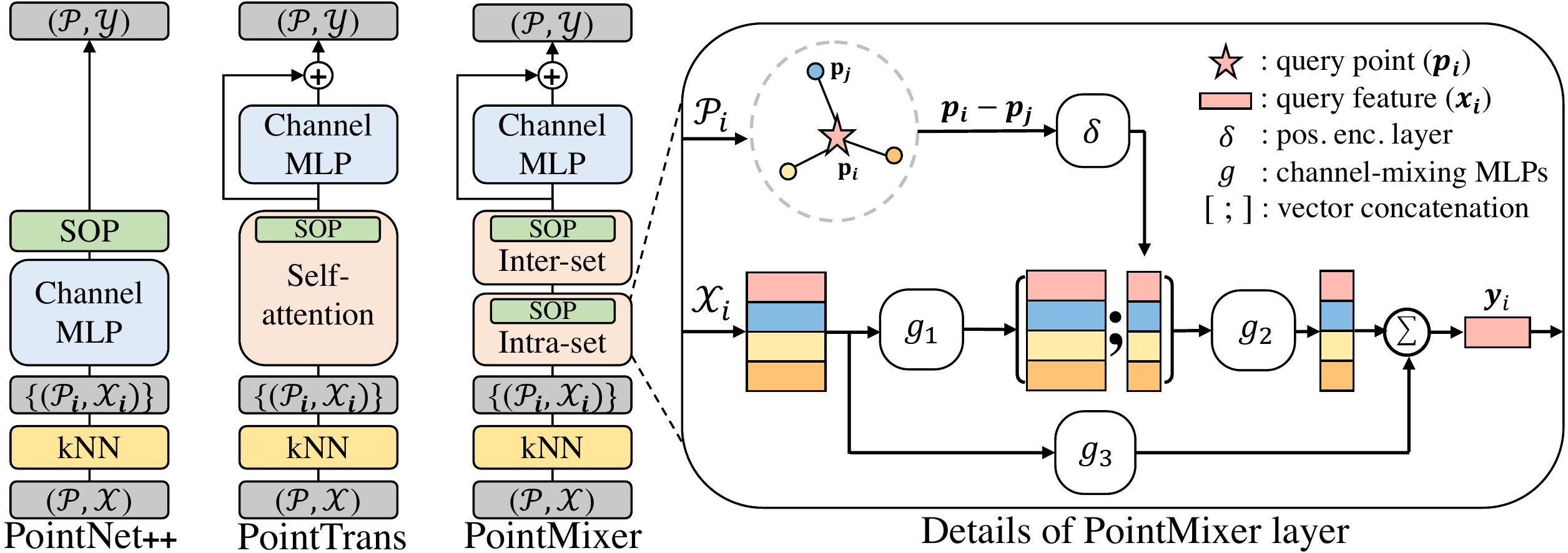}
\vspace{-2mm}
\caption{Block design comparison and PointMixer layer details. Note that SOP means symmetric operation, such as pooling and summation. 
}
\label{fig:fig_block}
\vspace{-4mm}
\end{figure}

\subsection{Preliminary}
\label{subsec:preliminary}
Let's assume a point cloud~$\mathcal{P}=\{\mathbf{p}_i\}_{i=1}^N$ and its corresponding point features~$\mathcal{X}=\{\mathbf{x}_i\}_{i=1}^N$ where $\mathbf{p}_i\in\mathbb{R}^3$ is the position of the $i$-th point and $\mathbf{x}_i\in\mathbb{R}^C$ is its corresponding feature. The objective is to learn a function $f:\mathcal{X}\rightarrow\mathcal{Y}$ to produce the output point features~$\mathcal{Y}=\{\mathbf{y}_i\}_{i=1}^N$. Instead of processing the entire point cloud directly, most approaches treat data locally based on points' proximity. For this purpose, $k$-Nearest Neighbor ($k$NN)~\cite{flann,faiss,open3d} is widely used to get an index map of neighbor points, which is denoted as:
\begin{align}
    \mathcal{M}_i = k\text{NN}(\mathcal{P},\mathbf{p}_i), 
    \label{eq:knn}
\end{align}
where $\mathcal{M}_i$ is an index map of the $K$ closest points for a query point~$\mathbf{p}_i\in\mathbb{R}^{3}$.
%
%
In other words, given a query index~$i$, the index map~$\mathcal{M}_i$ is defined as a set of $K$ nearest neighbor indices ($K=|\mathcal{M}_i|$). Accordingly, $k$NN can be understood as a directional graph that links each query point $\mathbf{p}_i$ to its $K$ closest neighbor points $\mathcal{P}_i=\{\mathbf{p}_j\in\mathcal{P}|j\in\mathcal{M}_i\}$ and the corresponding features $\mathcal{X}_i=\{\mathbf{x}_j\in\mathcal{X}|j\in\mathcal{M}_i\}$.

\noindent \textbf{PointNet\plusplus}~\cite{pointnet++} addresses the problem of PointNet~\cite{pointnet} that has difficulty in capturing local responses. 
To cope with this problem, it utilizes $k$NN and Farthest Point Sampling algorithm and builds asymmetric encoder-decoder network. Instead of dealing with the entire point cloud directly, PointNet\plusplus aggregates set-level responses locally as follows:
\vspace{-2mm}
\begin{equation}
\mathbf{y}_i = \operatorname{maxpool}_{j\in\mathcal{M}_i}\Big(\operatorname{MLP}\big([\mathbf{x}_j;\mathbf{p}_i-\mathbf{p}_j]\big)\Big),
\label{eq:pointnet++-mlp}
\vspace{-1mm}
\end{equation}
where $\mathbf{y}_i$ is the output feature vector for the $i$-th point and $[;]$ denotes vector concatenation.
By adopting this grouping and sampling strategy, MLPs can capture local responses from unordered 3D points.

\noindent \textbf{Point Transformer}~\cite{point-transformer} adopts vector subtraction attention as a similarity measurement for token-wise communication. Following PointNet\plusplus, Point Transformer also uses $k$NN to compute local responses. Given a query point feature~$\mathbf{x}_i$ and its neighbor feature set~$\mathcal{X}_i$, Point Transformer operates self-attention layers to densely relate token interaction as:
%
%
\vspace{-2mm}
\begin{equation}
\mathbf{y}_i = \sum_{j \in \mathcal{M}_i} \operatorname{softmax}\Big(\psi\big(\mathbf{W}_1\mathbf{x}_i-\mathbf{W}_2\mathbf{x}_j+\delta(\mathbf{p}_i-\mathbf{p}_j)\big)\Big)\Big(\mathbf{W}_3\mathbf{x}_j+\delta(\mathbf{p}_i-\mathbf{p}_j)\Big),
\label{eq:point_transformer}
\end{equation}
where $\mathbf{W}$ indicates a linear transformation matrix, $\psi(\cdot)$ denotes an MLPs to calculate vector similarities, 
$\delta(\mathbf{p}_i-\mathbf{p}_j)$ is a positional encoding vector to embed local structures of 3D points,
and $\mathbf{p}_i-\mathbf{p}_j$ is the relative distance between a query point~$\mathbf{p}_i$ and its neighbor point~$\mathbf{p}_j$. 

\subsection{MLP-Mixer as Point Set Operator}
\label{subsec:MLP-Mixer as Point Set Operator}
MLP-Mixer~\cite{mlp-mixer} has achieved remarkable success by only using MLPs for image classification. However, when dealing with sparse and unordered points, the direct application of the MLP-Mixer network is restricted. Let us revisit MLP-Mixer to ease the understanding of our PointMixer.

The MLP-Mixer layer\footnote{We set the relation of terminologies as $\text{layer}\subset\text{block}\subset\text{network}$.} consists of token-mixing MLPs and channel-mixing MLPs. MLP-Mixer takes $K$ tokens having $C$-dimensional features, denoted as $\mathbf{X}\in\mathbb{R}^{K\times C}$, where tokens are features from image patches. It begins with token-mixing MLPs that transposes the spatial axis and channel axis to mix spatial information. Then, it continues with channel-mixing MLPs so that input tokens are mixed in spatial and channel dimensions. 
\begin{equation}
\mathbf{X}' = \mathbf{X} + (\mathbf{W}_\text{2} \ \rho(\mathbf{W}_\text{1} (\operatorname{LayerNorm}(\mathbf{X}))^{\top}))^{\top},
\label{eq:mix_spatial}
\end{equation}
\begin{equation}
\mathbf{Y} = \mathbf{X}' + \mathbf{W}_\text{4} \ \rho(\mathbf{W}_\text{3} \operatorname{LayerNorm}(\mathbf{X}')),
\label{eq:mix_channel}
\end{equation}
where $\mathbf{W}$ is the weight matrix of a linear function and $\rho$ is $\operatorname{GELU}$~\cite{GELUs}. By \Eref{eq:mix_spatial}, token-mixing MLPs are sensitive to the order of the input tokens, which is permutation-variant property. Due to its property, positional encoding is not required in the vanilla MLP-Mixer as stated in the paper~\cite{mlp-mixer}.
%
%
However, this property is not desirable for processing irregular and unordered point clouds, which is different characteristics of uniform and ordered pixels in the image. To cope with this issue, previous point-based layers~\cite{pointnet++,point-transformer} are independent to orders of input point, \ie, permutation-invariance\footnote{To deal with unordered points, layers are permutation-invariant ($f_\text{layer}:\mathcal{X}_i\rightarrow\mathbf{y}_i$) and blocks are permutation-equivariant ($f_\text{block}:\mathcal{X}\rightarrow\mathcal{Y}$).}.
%
%
Moreover, as a point set operator, vanilla MLP-Mixer layer only computes \emph{intra-set} relations as PointNet\plusplus and Point Transformer do. From this analysis, we observe room for improvement in point cloud understanding. We propose PointMixer layer that is permutation-invariant and can also be used for a learnable upsampling in \Sref{subsec:Point Mixer Layer}.

\subsection{PointMixer: Universal Point Set Operator}
\label{subsec:Point Mixer Layer}
We introduce an approach to embed geometric relations between points' features into the MLP-Mixer's framework. 
As illustrated in~\Fref{fig:fig_block}, the PointMixer layer takes a point set~$\mathcal{P}_i=\{\mathbf{p}_j\}$ and its associated point features set~$\mathcal{X}_i=\{\mathbf{x}_j\}$ as inputs in order to compute the output feature vector $\mathbf{y}_i$.
For a point $\mathbf{p}_i$, we first compute a score vector $\mathbf{s}=[s_1,...,s_K]$ to aggregate $\mathcal{X}_i$ as follows:
%
%
\begin{equation}
s_j = g_2\Big(\big[g_1(\mathbf{x}_j); \delta(\mathbf{p}_i-\mathbf{p}_j)\big]\Big)~~\text{where}~~\forall j\in\mathcal{M}_i,
\label{eq:set_attention_1}
\end{equation}
%
%
where $g(\cdot)$ is the channel-mixing MLPs, $\delta(\cdot)$ is the positional encoding MLPs, and $\mathbf{x}_j$ is a $j$-th element of the feature vector set~$\mathcal{X}_i$.
%
%
Note that we follow the relative positional encoding scheme~\cite{pointnet++,point-transformer} to deal with unstructured 3D points.
As a result, we obtain the score vector $\mathbf{s}\in\mathbb{R}^K$. Finally, the features of the $K$ adjacent points are gathered to produce a new feature vector~$\mathbf{y}_i$ as: 
%
%
\vspace{-2mm}
\begin{equation}
\mathbf{y}_i=\sum_{j\in\mathcal{M}_i}\operatorname{softmax}(s_j)\odot g_3 (\mathbf{x}_j),
\label{eq:set_attention_2}
\vspace{-2mm}
\end{equation} 
where $\operatorname{softmax}(\cdot)$ is the Softmax function that normalizes the spatial dimension, and $\odot$ indicates an element-wise multiplication.
%
%
Note that this symmetric operation (SOP) in the proposed PointMixer is different from both the average pooling in MLP-Mixer and the max pooling in PointNet\plusplus, as described in~\Tref{table:compare}.
%
%

As a set operator, PointMixer layer has different characteristics compared to both MLP-Mixer layer~\cite{mlp-mixer}. First, PointMixer layer sees relative positional information~$\delta(\mathbf{p}_i-\mathbf{p}_j)$ to encode the local structure of a point set.
Second, the vanilla MLP-Mixer layer does not have the Softmax function.
Last, PointMixer layer does not have token-mixing MLPs for scalability to arbitrary number of neighbor points and for permutation-invariance to deal with unordered points. 
Let us explain the reasons behind these differences.

\noindent \textbf{No token-mixing MLPs.} \
There are two reasons that we do not put token-mixing MLPs into PointMixer layer.
First, token-mixing MLPs are permutation-variant, which makes it incompatible with unordered point clouds. 
As stated in Point Transformer~\cite{point-transformer}, permutation-invariant property is a necessary condition, also for PointMixer layer. 
Second, while a given pixel in an image systematically admits 8 adjacent pixels, each 3D point does not have pre-defined number of neighbors, which is determined by the clustering algorithms, such as $k$NN~\cite{edgeconv,pointnet++,point-transformer}, radius neighborhood~\cite{kpconv}, or hash table~\cite{mink,point-recon}. 
Since token-mixing MLPs can only take fixed number of input points\footnote{Please refer to the supplementary material for further details.}, token-mixing MLPs are not suitable for handling various cardinality of point sets.
%
%

\begin{figure*}[!t]
\centering
\includegraphics[width=1.00\linewidth]{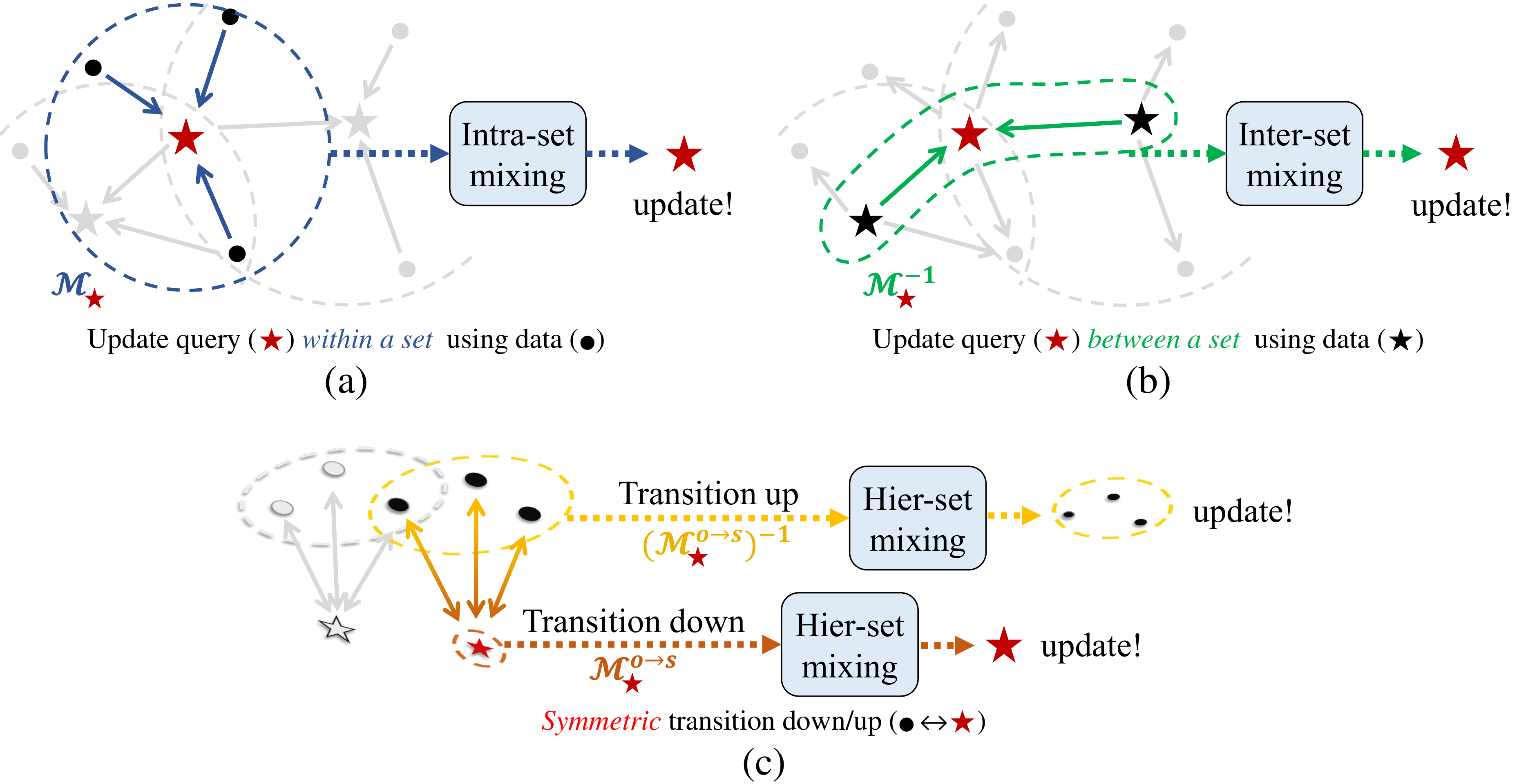}
\vspace{-6mm}
\caption{PointMixer is a universal point set operator: (a)~intra-set mixing, (b)~inter-set mixing, and (c)~hierarchical-set mixing.}
\vspace{-2mm}
\label{fig:fig_operator}
\end{figure*}

Inspired by Synthesizer~\cite{synthesizer}, we alleviate this problem by replacing the token-mixing MLPs with the Softmax function. We conjecture that the Softmax function weakly binds token-wise information in a non-parametric manner (\ie, normalization). By doing so, PointMixer layer can calculate arbitrary cardinality of point sets and have a permutation-invariant property. 
As described in~\Fref{fig:fig_operator} and in~\Tref{table:compare}, PointMixer layer can be used as a universal point operator for mixing various types of point sets: intra-set, intra-set, and hierarchical-set.

\noindent \textbf{Intra-set mixing.} \
Given a point set~$\mathcal{P}_i$ and its corresponding feature set~$\mathcal{X}_i$, intra-set mixing aims to compute point-wise interaction within each set. 
Usually, $k$NN is widely used to cluster neighbor points into groups. 
For example in~\Fref{fig:fig_operator}-(a), we apply $k$NN on each query point where $K{=}3$. As a result, a red point~($\textcolor{red}{\bigstar}$) has three neighbor points ($\textcolor{black}{\newmoon}$).
Based on the index map~$\mathcal{M}_i$ from $k$NN, the PointMixer layer updates a query feature~$\mathbf{x}_i$ using its neighbor feature set~$\mathcal{X}_i$, as in~\Eref{eq:set_attention_1}~and~\Eref{eq:set_attention_2}. While intra-set mixing is proven to be useful in various methods~\cite{point-cloud-analysis}, the receptive field is bounded within a set, as depicted in~\Fref{fig:fig_operator}-(a). To overcome this restriction, we propose the inter-set mixing operation.

\noindent \textbf{Inter-set mixing.} \
This is a new concept of spreading point features between different sets. Using the index mapping~$\mathcal{M}_i$, we can trace back to find another set~$\mathcal{P}_j$ that includes a query point~$\mathbf{p}_i$ as their neighbors. This process can be viewed as the inverse version of $k$NN, and we define the inverse mapping of $\mathcal{M}_i$ as $\mathcal{M}_{i}^{-1}=\{j|i\in\mathcal{M}_j\}$. 
For example in~\Fref{fig:fig_operator}-(b), given index mapping~$\mathcal{M}_{\textcolor{red}{\bigstar}}$, we compute inverse index mapping~$\mathcal{M}_{\textcolor{red}{\bigstar}}^{-1}$. Then, we can find the two adjacent sets whose query points are black points ($\textcolor{black}{\bigstar}$).
It implies the red point is included in two adjacent sets, as drawn in~\Fref{fig:fig_operator}-(\textcolor{black}{b})\footnote{There is a chance to collect variable number of points after an inverse mapping~$\mathcal{M}_{i}^{-1}$.}. In shorts, inverse mapping~$\mathcal{M}_{i}^{-1}$ finds the set index~$j$ that includes a point~$\mathbf{p}_i$. By doing so, inter-set mixing can aggregate point responses between neighbor sets~$\mathcal{P}_j$ into the query point $\mathbf{p}_i$.

\noindent \textbf{Hierarchical-set mixing.} \
The PointMixer layer is universally applicable for the transition down/up layers as shown in~\Fref{fig:fig_operator}-(c).
For instance, let's prepare a point set $\mathcal{P}_{s}=\{\mathbf{p}_j\}_{j=1}^{N'}$ that is sampled from original point set~$\mathcal{P}_{o}=\{\mathbf{p}_i\}_{i=1}^{N}$ ($\mathcal{P}_{s}\subset\mathcal{P}_{o}$).
Using a point $\mathbf{p}_j\in\mathcal{P}_{s}$, we calculate its neighbors from $\mathcal{P}_{o}$ and obtain index mapping $\mathcal{M}_j^{o\rightarrow s}$. By putting $\mathcal{M}_j^{o\rightarrow s}$ in~\Eref{eq:set_attention_1}~and~\Eref{eq:set_attention_2}, we readily pass the feature from $\mathcal{P}_{o}$ to $\mathcal{P}_{s}$ (\ie,~point~\emph{downsampling}), which is computed as:
%
%
\begin{equation}
\mathcal{M}_j^{o\rightarrow s} = k\text{NN}(\mathcal{P}_o,\mathbf{p}_j)~~\text{where}~~\forall \mathbf{p}_j\in\mathcal{P}_s.
\end{equation}

\begin{figure}[!t]
\centering
\includegraphics[width=0.85\linewidth]{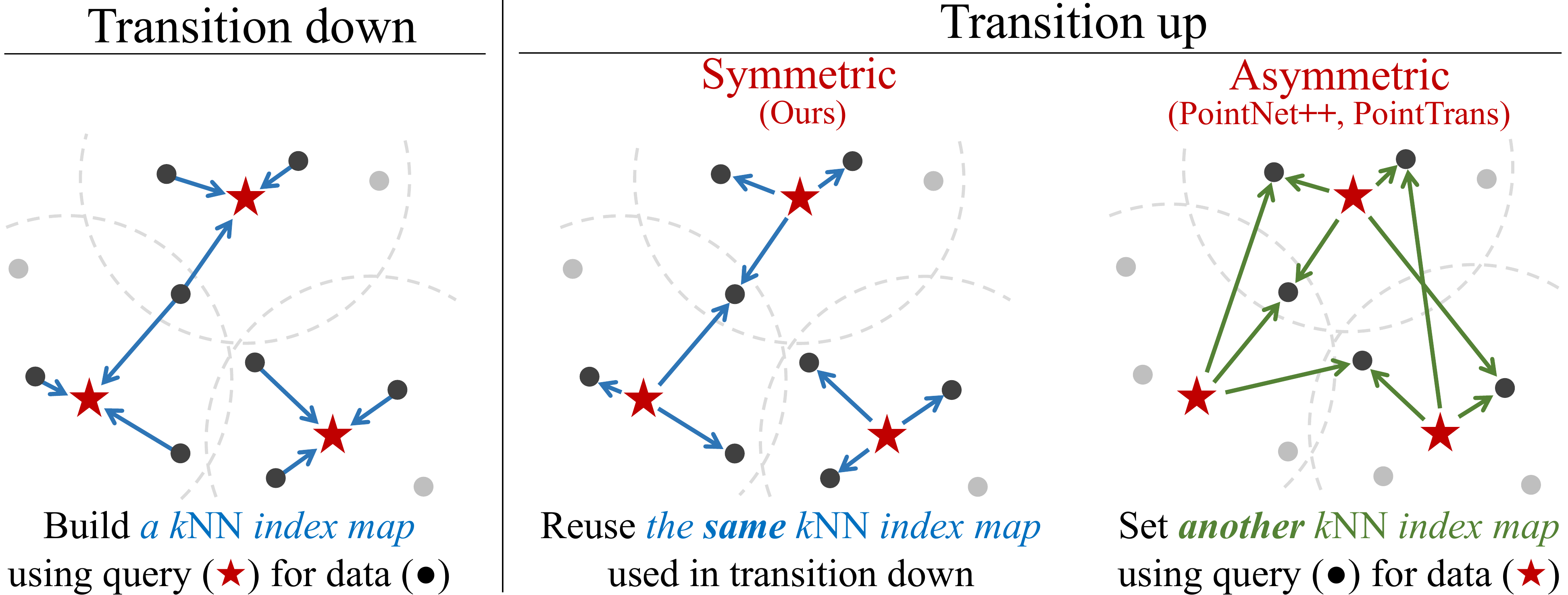}
\vspace{-2mm}
\caption{Comparison of transition layer: symmetric (ours), asymmetric~\cite{pointnet++,zhao2019pointweb,point-transformer}.}
\label{fig:fig_symmetric}
\vspace{-2mm}
\end{figure}

For point~\emph{upsampling}, we notice that conventional U-Net in both PointNet\plusplus and Point Transformer is \emph{not symmetric} in terms of downsampling and upsampling. This is because the spatial grouping is performed asymmetrically as visualized in~\Fref{fig:fig_symmetric}. In details, conventional approaches~\cite{pointnet++,point-transformer,zhao2019pointweb} build another $k$NN map~$\mathcal{M}_i^{s\rightarrow o}$ for the upsampling as below:
\begin{equation}
\mathcal{M}_i^{s\rightarrow o} = k\text{NN}(\mathcal{P}_s,\mathbf{p}_i)~~\text{where}~~\forall \mathbf{p}_i\in\mathcal{P}_o.
\end{equation}
However, this is not symmetric because \emph{nearest neighbor is not a symmetric function}: even if $\mathbf{p}_i$'s nearest neighbor is $\mathbf{p}_j$, $\mathbf{p}_j$'s nearest neighbor may not $\mathbf{p}_i$. 

Instead of creating a new index map~$\mathcal{M}_i^{s\rightarrow o}$, our PointMixer layer can use $\big(\mathcal{M}_j^{o\rightarrow s}\big)^{{-}1}$ for upsampling (\Fref{fig:fig_operator}-(\textcolor{black}{b})) by re-using the original index mapping from the downsampling~$\mathcal{M}_j^{o\rightarrow s}$.
See \Fref{fig:fig_symmetric} for the symmetric upsampling. The benefit of this approach is that it does not introduce additional $k$-Nearest Neighbor search. 
Furthermore, we can propagate point responses in different hierarchy based on their scores computed by our PointMixer, instead of using trilinear interpolation~\cite{pointnet++,point-transformer}. These technical differences results in higher performance than that of asymmetric design on dense prediction tasks (see~\Sref{subsec:Ablation Study}).

As illustrated in~\Fref{fig:fig_arch}-(\textcolor{black}{b}) of the transition down block, we use Farthest Point Sampling algorithm to produce $\mathcal{P}_s$ from $\mathcal{P}_o$. Then, we utilize $k$NN to sample $\mathcal{P}_s$ from $\mathcal{P}_o$. The resulting point locations are used to calculate the relative distance~$\mathbf{p}_i-\mathbf{p}_j$. In the transition up block, we keep using the index map~$\mathcal{M}_j^{o\rightarrow s}$ calculated in the transition down block. To this end, we apply the PointMixer layer in transition up/down while maintaining the symmetric relation between the sampled point cloud~$\mathcal{P}_s$ and the original points~$\mathcal{P}_o$. We empirically prove that this symmetric upsampling layer helps the network to predict dense point-level representations accurately in~\Sref{sec:Experiments}.
%

\subsection{Network Architecture}
\label{subsec:Network Architecture}
\begin{figure*}[!t]
\centering
\includegraphics[width=1.00\linewidth]{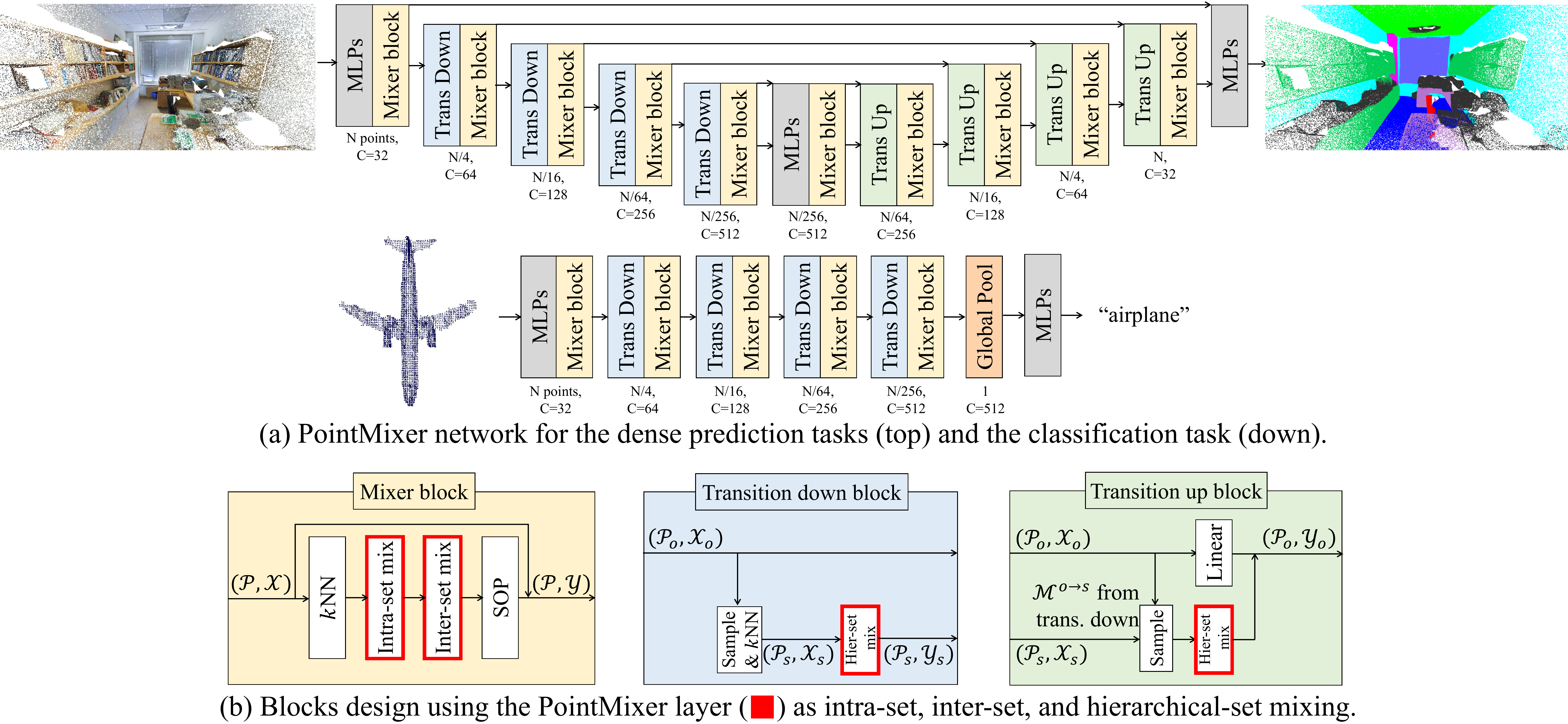}
\vspace{-6mm}
\caption{Overall architecture of PointMixer network.}
\vspace{-2mm}
\label{fig:fig_arch}
\end{figure*}

In this section, we describe the details of our MLP-like encoder-decoder architecture as shown in~\Fref{fig:fig_arch}.
Our network is composed of several MLP blocks, such as the transition down blocks, transition up blocks, and Mixer blocks. For a fair comparison, our network mainly follows the proposed hyper-parameters in Point Transformer~\cite{point-transformer} for network composition. Overall, our network takes a deep pyramid-style network that progressively downsamples points to obtain global features. For dense prediction tasks, such as semantic segmentation or point reconstruction, we include upsampling blocks for per-point estimation. Finally, our header block is designed for task-specific solutions. 
For classification, we take fully-connected layers, dropout layers, and global pooling layer. For semantic segmentation and point reconstruction, the header block consists of MLPs without pooling layers.


\section{Experiments}
\label{sec:Experiments}
\vspace{-1mm}
In this section, we evaluate the efficacy and versatility of the proposed PointMixer for various point cloud understanding tasks: semantic segmentation~\cite{armeni_cvpr16}, point reconstruction~\cite{point-recon}, and object classification~\cite{modelnet40}. 

\vspace{-1mm}
\subsection{Semantic Segmentation}
\label{subsec:Semantic segmentation}
To validate the effectiveness of PointMixer for semantic segmentation, we propose an evaluation on the large-scale point cloud dataset S3DIS~\cite{armeni_cvpr16} consisting of 271 room reconstructions. Each 3D point of this dataset is assigned to one label among 13 semantic categories.
For the sake of fairness, we meticulously follow the widely used evaluation protocol proposed by Point Transformer~\cite{point-transformer}. For training, we set the batch size as $4$ and use the SGD optimizer with momentum and weight decay set to $0.9$ and $0.0001$ respectively. 
For evaluation, we use the class-wise mean Intersection of Union (mIoU), class-wise mean accuracy (mAcc), and overall point-wise accuracy (OA).

\begin{figure*}[!t]
\centering
\includegraphics[width=0.97\linewidth]{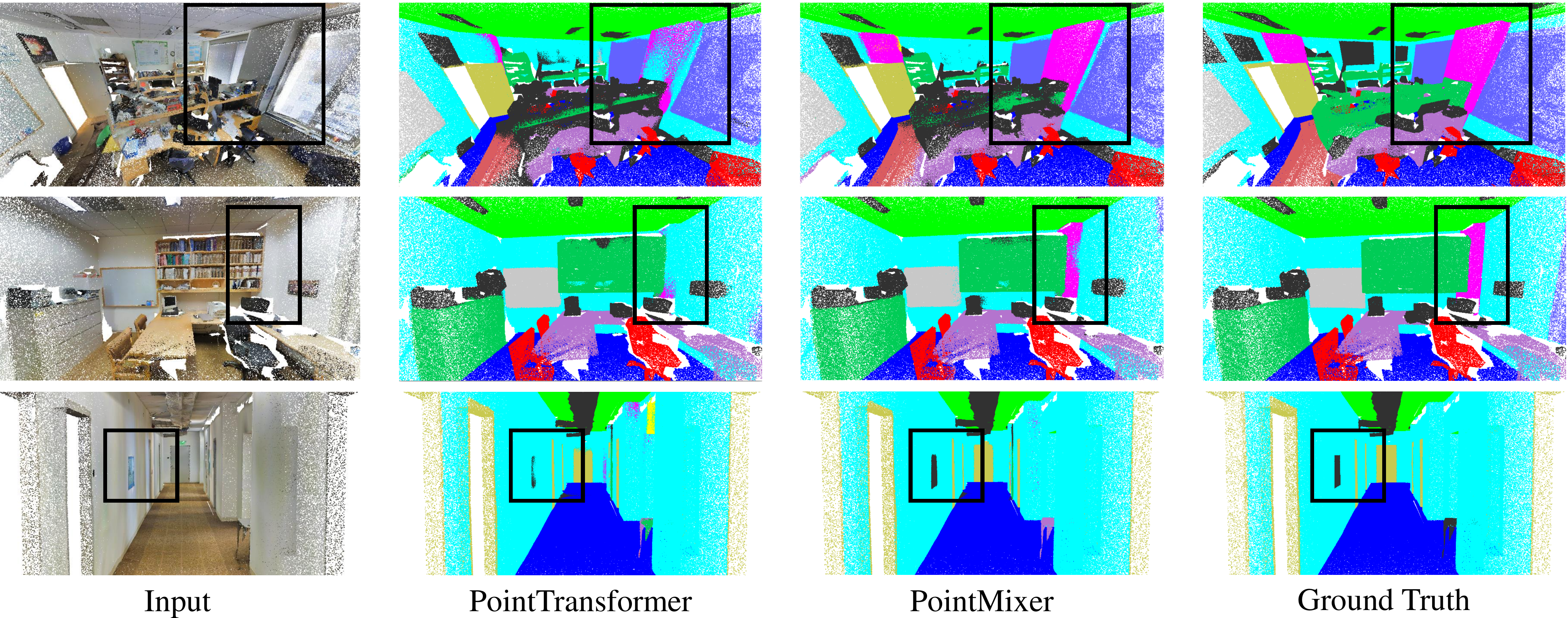}
\vspace{-4mm}
\caption{Qualitative results in semantic segmentation on S3DIS Area 5 test dataset~\cite{armeni_cvpr16}.}
\vspace{-4mm}
\label{fig:fig_qual_semseg}
\end{figure*}

As shown in~\Tref{table:semseg} and~\Fref{fig:fig_qual_semseg}, PointMixer achieves the state-of-the-art performance in \textsc{S3DIS}~\cite{armeni_cvpr16} Area 5, though PointMixer network consumes less parameters (\textbf{6.5M}) than that of Point Transformer (7.8M). Even in class-wise IoU, PointMixer network outperforms Point Transformer~\cite{point-transformer}, except for a few classes. 
%
%
While various studies~\cite{point-transformer,cloud-transformer,point-cloud-transformer} underline the necessity of dense point-wise interaction (\ie, self-attention layer), PointMixer successfully outperforms these approaches purely using Channel MLPs. These results consistently support that dense token communication is not an essential factor as stated in Synthesizer~\cite{synthesizer} and Metaformer~\cite{metaformer}. We claim that it is much more crucial to mix information through various point sets.
%
%
Moreover, the experimental result shows that our symmetric upsampling layer is more effective for semantic segmentation than heuristic sampling-based asymmetric upsampling layer which all of previous approaches~\cite{pointnet++,kpconv,point-transformer} have used.
We further discuss the effectiveness of our hierarchical-set mixing layer with ablation studies in \Sref{subsec:Ablation Study}. 
\begin{table*}[!t]
\setlength{\tabcolsep}{2pt}
\caption{Semantic segmentation results on S3DIS Area 5 test dataset~\cite{armeni_cvpr16}.
}
\vspace{-3.5mm}
\resizebox{1.00\linewidth}{!}{
\begin{tabular}{llrcccccccccccccccc}
\Xhline{4\arrayrulewidth}

Method & &
Param. & mAcc & mIoU & & 
ceiling & floor & wall & column & window & door & table & chair & sofa & book. & board & clutter \\ \Xhline{2\arrayrulewidth}



PAConv~\cite{xu2021paconv} & &
- & 73.0 & 66.6 & & 
\first{94.6} & \second{98.6} & 82.4 & 26.4 & 58.0 & 60.0 & 80.4 & 89.7 & 69.8 & 74.3 & 73.5 & 57.7 \\ 

KPConv \textit{deform}~\cite{kpconv} & &
- & 72.8 & 67.1 & & 
92.8 & 97.3 & 82.4 & 23.9 & 58.0 & 69.0 & 81.5 & \first{91.0} & \first{75.4} & 75.3 & 66.7 & 58.9 \\  

MinkowskiNet~\cite{mink} & &
37.9M & 71.7 & 65.4 & & 
91.8 & \first{98.7} & \second{86.2} & 34.1 & 48.9 & 62.4 & 81.6 & \second{89.8} & 47.2 & 74.9 & 74.4 & 58.6 \\



PointTrans~\cite{point-transformer} & &
\second{7.8M} & 76.5 & \second{70.4} & & 
94.0 & 98.5 & \first{86.3} & 38.0 & \first{63.4} & 74.3 & \second{89.1} & 82.4 & \second{74.3} & \first{80.2} & 76.0 & \second{59.3} \\ 

FastPointTrans~\cite{park2022fast} & &
37.9M & \first{77.9} & 70.3 & & 
\second{94.2} & 98.0 & 86.0 & \first{53.8} & 61.2 & \second{77.3} & 81.3 & 89.4 & 60.1 & 72.8 & \first{80.4} & 58.9 \\ \hline 

PointMixer (ours) & &
\first{6.5M} & \second{77.4} & \first{71.4} & & 
\second{94.2} & 98.2 & 86.0 & \second{43.8} & \second{62.1} & \first{78.5} & \first{90.6} & 82.2 & 73.9 & \second{79.8} & \second{78.5} & \first{59.4} \\ \Xhline{4\arrayrulewidth}

\end{tabular}
}
\vspace{-2mm}
\label{table:semseg}
\end{table*}


\subsection{Point Cloud Reconstruction}
\label{subsec:Point Cloud Reconstruction}
To highlight the versatility of our approach, we propose a large-scale assessment for the newly introduced task of point cloud reconstruction~\cite{point-recon} where an inaccurate point cloud is jointly denoised, densified, and completed. This experiment is also particularly interesting to evaluate the generalization capabilities of networks since it allows us to test methods in unmet environments. 
Specifically, we train on the synthetic objects of ShapeNetPart~\cite{shapenet-part} and evaluate the reconstruction accuracy on on unmet indoor scenes from ScanNet~\cite{scannet} (real reconstruction) and ICL-NUIM~\cite{icl-nuim} (synthetic data).
Note that in~\cite{point-recon}, the point reconstruction is performed in two stages: 1) point upsampling via a sparse hourglass network, 2) denoising and refinement via Transformer network. For this evaluation, we propose to replace the second stage of this pipeline with various architectures (\ie, PointMixer network and previous studies~\cite{pointnet,pointnet++,point-transformer}) to compare their performances. Under the same data augmentation and data pre-processing as in~\cite{point-recon,pcu-dispu}, we train PointMixer and previous studies~\cite{pointnet,pointnet++,point-transformer}. For evaluation, we utilize the Chamfer distance (CD) to measure the distance between the predictions and the ground truth point clouds. Additionally, we use the accuracy (Acc.), completeness (Cp.), and F1 score to measure the performance in occupancy aspects~\cite{dtu_dataset_0,dtu_dataset_1,mvsnet,tanks-and-temples}.

\begin{figure}[!t]
\centering
\includegraphics[width=1.00\linewidth]{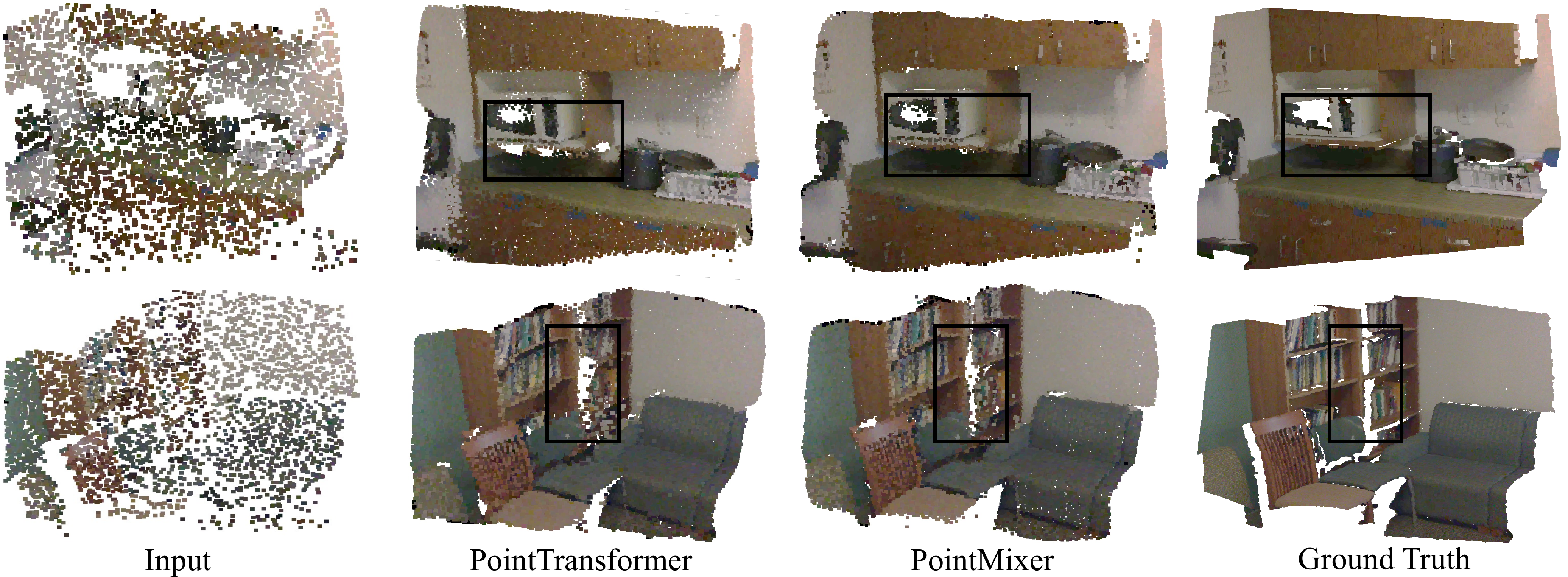}
\vspace{-6mm}
\caption{Qualitative results in point reconstruction on ScanNet dataset~\cite{scannet}.
}
\vspace{-4mm}
\label{fig:fig_qual_recon}
\end{figure}

Though our network is solely trained in a synthetic/object dataset, our network can generalize towards unmet scenes including real-world 3D scans~\cite{scannet} and synthetic/room-scale point clouds~\cite{icl-nuim} as in~\Tref{table:recon} and ~\Fref{fig:fig_qual_recon}. 
Moreover, our method compares favorably to previous Transformer-based~\cite{point-transformer,point-recon} and MLP-based~\cite{pointnet,pointnet++} studies. In particular, the performance gap between ours and previous studies become larger in the ScanNet~\cite{scannet} and ICL-NUIM dataset~\cite{icl-nuim}, which indicates better generalization performance.

\begin{table*}[!t]
\setlength{\tabcolsep}{2pt}
\caption{Point cloud reconstruction results.
}
\vspace{-3mm}
\resizebox{1.00\linewidth}{!}{
\begin{tabular}{llcccclcccclcccc}
\Xhline{4\arrayrulewidth}

\multirow{2}{*}{Method} & &
\multicolumn{4}{c}{ShapeNet-Part~\cite{shapenet-part}} & &
\multicolumn{4}{c}{ScanNet~\cite{scannet}} & &
\multicolumn{4}{c}{ICL-NUIM~\cite{icl-nuim}} 
\\ \cline{3-6} \cline{8-11} \cline{13-16}

& &
CD($\downarrow$) & Acc.($\uparrow$) & Cp.($\uparrow$) & F1($\uparrow$) & &
CD($\downarrow$) & Acc.($\uparrow$) & Cp.($\uparrow$) & F1($\uparrow$) & &
CD($\downarrow$) & Acc.($\uparrow$) & Cp.($\uparrow$) & F1($\uparrow$)
\\ \Xhline{2\arrayrulewidth}


PointNet~\cite{pointnet} & &
1.33 & 63.2 & 38.8 & 48.2 & &
3.05 & 37.5 & 27.8 & 32.6 & &
2.98 & 46.9 & 33.2 & 38.1
\\

PointNet\plusplus~\cite{pointnet++} & &
1.25 & 65.1 & 39.0 & 50.1 & &
2.97 & 38.3 & 29.5 & 33.4 & &
2.88 & 48.8 & 35.8 & 39.9
\\

PointRecon~\cite{point-recon} & &
1.19 & \first{81.0} & 40.4 & \second{53.4} & &
2.86 & 40.4 & 30.2 & 34.1 & &
2.78 & \second{54.1} & \second{38.1} & \second{43.6}
\\

PointTrans~\cite{point-transformer} & &
\second{1.12} & 75.9 & \second{40.9} & 52.7 & &
\second{2.79} & \second{41.1} & \second{32.1} & \second{35.6} & &
\second{2.57} & {51.1} & {36.4} & {41.6}
\\ \hline 

PointMixer (ours) & &
\first{1.11} & \second{77.1} & \first{41.5} & \first{53.7} & &
\first{2.74} & \first{42.1} & \first{33.5} & \first{37.8} & &
\first{2.43} & \first{56.5} & \first{38.2} & \first{44.7}
\\ \Xhline{4\arrayrulewidth}

\end{tabular}
}
\label{table:recon}
\end{table*}

\subsection{Shape Classification}
\label{subsec:Shape Classification}
\begin{wraptable}{r}{0.40\textwidth}
\centering
\vspace{-12mm} 
\caption{Shape classification results on ModelNet40 dataset~\cite{modelnet40}. 
}
\vspace{0.5mm}
\resizebox{1.0\linewidth}{!}{
\begin{tabular}{llrcc}
\Xhline{4\arrayrulewidth}

Method & &
Param. & mAcc & OA
\\ \Xhline{2\arrayrulewidth}


PointNet~\cite{pointnet} & &
\multicolumn{1}{c}{-} & 86.2 & 89.2
\\

PointNet\plusplus~\cite{pointnet++} & &
\first{1.4M}& - & 90.7 
\\


PointConv~\cite{wu2019pointconv} & &
18.6M & - & 92.5
\\

DGCNN~\cite{edgeconv} & &
\multicolumn{1}{c}{-} & 90.2 & 92.9 
\\

KPConv \textit{rigid}~\cite{kpconv} & &
15.2M& - & 92.9
\\



PointTrans~\cite{point-transformer} & &
5.3M& \second{90.6} & \first{93.7}
\\ \hline 

PointMixer (ours) & &
\second{3.9M}& \first{91.4} & \second{93.6}
\\ \Xhline{4\arrayrulewidth}

\end{tabular}
}
\vspace{-8mm}
\label{table:class}
\end{wraptable}

The ModelNet40~\cite{modelnet40} dataset has 12K CAD models with 40 object categories. We follow the official train/test splits to train and evaluate ours and the previous studies. 
For fair comparison, we follow the data augmentation and pre-processing as proposed in PointNet\plusplus~\cite{pointnet++}, which is also adopted in the recent studies~\cite{point-transformer,interpconv,kpconv,xu2021paconv}. For evaluation, we adopt the mean accuracy within each category (mAcc), and the overall accuracy over all classes (OA) with the same evaluation protocol with previous approaches~\cite{pointnet++,point-transformer,xu2021paconv}.
%
%
The results presented in~\Tref{table:class} show that our approach outperforms the recently proposed techniques.
Especially, PointMixer achieves the highest mAcc with outperforming other methods by a large margin as 0.8 mAcc, using 1024 points without normals. Moreover, our network shows this competitive performance with less parameters (\textbf{3.9M}) against previous studies (Point Transformer\footnote{Since there is no official release of codes, we use the best implementation of Point Transformer available in the public domain, which contains the official code provided by the authors of Point Transformer and reproduces the reported accuracy.} 5.3M, KPConv 15.2M).
Based on these comparisons with Point Transformer and related work, we conclude that PointMixer network effectively and efficiently aggregates various point responses through intra-set mixing, inter-set mixing, and downsampling layers, even for shape classification task.

\subsection{Ablation Study}
\label{subsec:Ablation Study}
We conduct an extensive ablation study about the proposed PointMixer in semantic segmentation on the S3DIS dataset~\cite{armeni_cvpr16} and in point cloud reconstruction on the ShapeNet-Part dataset~\cite{shapenet-part}. 
First, we proceed the case study of our PointMixer~(\Tref{table:ablation-pe}, top-right one). 
Second, we analyze the influence of positional encoding on PointMixer (\Tref{table:ablation-pe}, left one). Last, we compare PointMixer that is free from token-mixing MLPs (\Tref{table:ablation-pe}, bottom-right one).

\noindent \textbf{Universal point set operator.} \
PointMixer can be applicable to act as intra-set mixing, inter-set mixing, and hierarchical-set mixing. Technically speacking, intra-set mixing and hierarchical-set mixing for point downsampling utilizes $\mathcal{M}_i$, but inter-set mixing and hierarchical-set mixing for point upsampling use $\mathcal{M}_i^{-1}$. This ablation study aims to validate the inverse mapping as well as a functionality of PointMixer layer. As in~\Tref{table:ablation-pe}, 
especially, when we combine the usage of hierarchical-set mixing and inter-set mixing, the synergy brings large performance improvement in both two tasks, 2.2 mIoU in semantic segmentation and 0.02 Chamfer distance in point reconstruction.
It implies that the various ways of sharing points' responses are beneficial for point cloud understanding.

\noindent \textbf{Unnecessary token-mixing MLPs.} \
We validate our claims about the role of the Softmax function, \ie, weakly binding tokens. For this purpose, we replace the Softmax functions with token-mixing MLPs, as proposed in the vanilla MLP-Mixers. \Tref{table:ablation-pe} demonstrates that even without explicit use of token-mixing MLPs, our PointMixer successfully achieves similar accuracy in semantic segmentation and point reconstruction. Moreover, it takes 3 days for training the PointMixer with token-mixing MLPs while our original PointMixer requires a day (twice faster) with less parameter consumption (18\% less). These results are consistently support to Metaformer~\cite{metaformer} and Synthesizer~\cite{synthesizer} in that dense token-wise interaction is not an essential choice.

\noindent \textbf{Positional encoding.} \
MLP-Mixer does not use positional encoding since token-mixing MLPs are sensitive to the order of tokens. However, relative position information is an important factor to handle unstructured 3D points. Also, our PointMixer layer is free from token-mixing MLPs to obtain permutation-invariant property to function as a universal point set operator. Without any modification on the layer except positional encoding layers~$\delta(\mathbf{p}_i-\mathbf{p}_j)$, we experiment the effectiveness of positional encoding in all of these usages. As in~\Tref{table:ablation-pe}, there is large performance drop when we intentionally omit positional encodings in our PointMixer layer. 
Different from MLP-Mixer, we claim the relative positional encoding is a necessary condition to deal with 3D points.

\noindent \textbf{Symmetric architecture.} \
When we think of the convolutional neural networks for image recognition, it is natural to design symmetric encoder network and decoder network. For instance in the previous paper~\cite{deconvnet}, this paper verify the importance of symmetric transposed convolution layer design for semantic segmentation task. In contrast, point-based studies~\cite{point-transformer,zhao2019pointweb} dominantly rely on the asymmetric PointNet\plusplus architecture, which is similar to FCN~\cite{long2015fully}.

The top-right sub-table in~\Tref{table:ablation-pe} supports our claim. When we do not use hierarchical-set mixing, the architecture become asymmetric, and transition down and up layers are identical to that in PointNet\plusplus~\cite{pointnet++} and Point Transformer~\cite{point-transformer}. It turns out that the asymmetric architecture (the second row) degrades the performance of the network (the last row) in terms of both mIoU and Chamfer distance by $2.3$ absolute percentage and $0.01$, respectively. Moreover, when we apply inter-set mixing, the performance gap further increases in both semantic segmentation and point reconstruction tasks.
\begin{table}[!t]
\centering
\vspace{1mm}
\caption{Ablation study about positional encoding~(left), types of mixing~(top-right) and token-mixing MLPs~(bottom-right). In the bottom-right table, we denote channel-mixing MLPs and token-mixing MLPs as C-MLP and T-MLP, respectively. Note that number of parameters (Param.) and training time (Time) is measured under semantic segmentation task on S3DIS dataset~\cite{armeni_cvpr16}.}
\vspace{-3mm}
\resizebox{0.492\linewidth}{!}{
\setlength{\tabcolsep}{2pt}
\begin{tabular}{ccclcclcc}
\Xhline{4\arrayrulewidth}

\multicolumn{3}{c}{Preserve (\checkmark)} & &
\multicolumn{2}{c}{Semantic seg} & &
\multicolumn{2}{c}{Point recon} 
\\ \cline{1-3} \cline{5-6} \cline{8-9} 

\multicolumn{3}{c}{Pos. enc. in} & &  
\multirow{2}{*}{mIoU} & \multirow{2}{*}{mAcc} & &
\multirow{2}{*}{CD($\downarrow$)} & \multirow{2}{*}{~~F1($\uparrow$)~~} 
\\ 

~Intra~ & ~Inter~ & ~Hier~ & &
& & &
& 
\\ \Xhline{2\arrayrulewidth}

& & & & 
64.7 & 71.8 & & 
1.13 & 53.0
\\ 

& & \checkmark & &
65.9 & 72.5 & &
1.13 & 53.0
\\ 

& \checkmark & & &
64.0 & 70.5 &  & 
1.13 & 53.0
\\ 

& \checkmark & \checkmark & &
66.1 & 72.9 & 
& \second{1.12} & \second{53.6}
\\ \Xhline{2\arrayrulewidth}

\checkmark & & & &
69.6 & 76.3 &  & 
\second{1.12} & \second{53.6}
\\ 

\checkmark & & \checkmark & &
\second{70.2} & \second{76.8}  & 
& 1.13 & 52.5
\\ 

\checkmark & \checkmark & & &
69.9 & 76.5 & &
1.13 & 53.5 
\\ 

\checkmark & \checkmark & \checkmark & &
\first{71.4} & \first{77.4} & &
\first{1.11} & \first{53.7}
\\ \Xhline{4\arrayrulewidth}

\end{tabular}} 
\resizebox{0.492\linewidth}{!}{ 
\setlength{\tabcolsep}{2pt}
\begin{tabular}{ccclcclcc}
\Xhline{4\arrayrulewidth}

\multicolumn{3}{c}{Preserve (\checkmark)} & &
\multicolumn{2}{c}{Semantic seg} & &
\multicolumn{2}{c}{Point recon} 
\\ \cline{1-3} \cline{5-6} \cline{8-9} 


~Intra~ & ~Inter~ & ~Hier~ & &
mIoU & mAcc & &
CD($\downarrow$) & F1($\uparrow$)
\\ \Xhline{2\arrayrulewidth}

\checkmark & & & &
69.2 & 75.8 & &
1.13 & \second{53.5} 
\\ 

\checkmark & \checkmark & & &
69.1 & 75.7 & &
\second{1.12} & \second{53.5}
\\ 

\checkmark & & \checkmark & &
\second{69.3} & \second{76.6} & &
\first{1.11} & \second{53.5} 
\\ 

\checkmark & \checkmark & \checkmark & &
\first{71.4} & \first{77.4} & &
\first{1.11} & \first{53.7}
\\ \Xhline{4\arrayrulewidth}

\vspace{-3mm}

\\ \Xhline{4\arrayrulewidth}


C-MLP & T-MLP & Softmax & &
Param. & Time & & 
mIoU & 
CD($\downarrow$)
\\ \Xhline{2\arrayrulewidth}

\checkmark & & & &
\first{6.5M} & \first{40h} & & 
58.3 & 1.23
\\

\checkmark & & \checkmark & &
\first{6.5M} & \second{44h} & & 
\first{71.4} & \first{1.11}
\\ \hline

\checkmark & \checkmark & & &
\second{7.4M} & 88h & & 
\second{71.1} & \second{1.12}
\\

\checkmark & \checkmark & \checkmark & &
\second{7.4M} & 95h & & 
\second{71.1} & \first{1.11}
\\ \Xhline{4\arrayrulewidth}

\end{tabular}} 
\label{table:ablation-pe}
\vspace{-2mm}
\end{table}

\section{Conclusion}
\label{sec:Conclusion}
We propose a MLP-like architecture for point cloud understanding, which focuses on sharing point responses in numerous and diverse point sets through a universal point set operator, \emph{PointMixer}.
Regardless of the cardinality of a point set, PointMixer layer can ``mix" point responses in intra-set, inter-set, and hierarchical-set. Moreover, we present a point-based general architecture that involves symmetric encoder-decoder blocks for propagating information through hierarchical point sets. 
Extensive experiments validate the efficacy of our PointMixer network with superior or compelling performance compared to Transformer-based studies. 
Through out this paper, we claim that dense token-wise calculation, such as self-attention layers or token-mixing MLPs, is not an essential choice for point cloud processing. Instead, we emphasize the importance of information sharing toward various point sets.

\section*{Acknowledgements}
(1) IITP grant funded by the Korea government(MSIT) (No.2019-0-01906, Artificial Intelligence Graduate School Program(POSTECH)) and (2) Institute of Information and communications Technology Planning and Evaluation (IITP) grant funded by the Korea government(MSIT) (No.2021-0-02068, Artificial Intelligence Innovation Hub)


\appendix
\section*{Supplementary Material}
This is a supplementary material for the paper, \textit{PointMixer: MLP-Mixer for Point Cloud Understanding}. We will further describe the details: 
efficiency analysis (\Sref{sec:efficiency}), 
%
%
point receptive fields comparison (\Sref{sec:receptive}), 
limited cardinality issues in token-mixing MLPs (\Sref{sec:tokenMLPs}), 
and task-specific training details~(\Sref{sec:Training details}).

\section{Efficiency analysis}
\label{sec:efficiency}
In this section, we analyze the latency and memory consumption of each point set layer: PointNet\plusplus layer~(\Eref{eq:pointnet++-mlp}), Point Transformer layer~(\Eref{eq:point_transformer}) and PointMixer layer~(\Eref{eq:set_attention_1} and \Eref{eq:set_attention_2}). We conduct this ablation study based on the PointMixer network for the 3D shape classification task. 
For a fair comparison, we \textbf{strictly maintain to use the same downsampling layers}.
Furthermore, we do not use inter-set mixing layer to keep the other components of the network the same.


We measure mAcc, OA, the average inference time per object, and the peak GPU memory usage of each method on ModelNet40 dataset~\cite{modelnet40}.
Note that we re-implement PointNet\plusplus\footnote{Since the original implementation of PointNet\plusplus~\cite{pointnet++} does not use the residual connection, this re-implementation brings performance gain to the model~\cite{pointnet++}.} and Point Transformer with the same number of residual blocks as our PointMixer, and train those models on the ModelNet40 dataset~\cite{modelnet40} with the same training configuration for a fair comparison.

\begin{table}[!h]
\centering
\caption{A comparison of efficiency.
}
\vspace{-3mm}
\resizebox{0.75\linewidth}{!}{
\begin{tabular}{lrrrcc}

\Xhline{4\arrayrulewidth} 

Layer type & Param. (M) & Memory (MB) & Latency (ms) & mAcc & OA
\\ \Xhline{2\arrayrulewidth} 

PointNet\plusplus~\cite{pointnet++} & \first{3.3} & \first{1463} & \first{13.04} & \second{90.9} & \second{93.2} 
\\ 

PointTrans~\cite{point-transformer} & 5.3 & 1473 & \second{19.77} & 90.2 & 93.1
\\ 

PointMixer (ours) & \second{3.5} & \second{1465} & 20.72 & \first{91.2} & \first{93.3}
\\ \Xhline{4\arrayrulewidth} 

\end{tabular}
}
\label{table:efficiency}
\end{table}




As shown \Tref{table:efficiency}, the network with Point Transformer layer~\cite{point-transformer} consumes the largest amount of GPU memory to infer a 3D object since it calculates a memory-consuming \textit{vector similarity}.
On the other hand, our PointMixer layer computes a \textit{scalar score}, denoted by $s_j$, to aggregate neighbor features, denoted by $\mathbf{x}_j$. It consequently consumes 8MB less GPU memory than Point Transformer layer~\cite{point-transformer} although both Point Transformer and Point Mixer layers are slower than PointNet\plusplus layer since both of them use the expensive $\operatorname{softmax}$ operation.
Furthermore, the PointMixer with only intra-set mixing outperforms PointNet\plusplus~\cite{pointnet++} layer by 0.3 mAcc and 0.1 OA although PointNet\plusplus~\cite{pointnet++} also requires much less memory than Point Transformer~\cite{point-transformer}.
This result implies that our score-based aggregation can embed local responses more effectively than simple pooling-based aggregation which PointNet\plusplus~\cite{pointnet++} uses.
As a result, our PointMixer layer can encode local relations within a point set both more effectively and efficiently than previous approaches~\cite{pointnet++,point-transformer}, along with the other strengths that inter-set mixing and hierarchical-mixing layers have, which are already shown in \Tref{table:semseg}, \Tref{table:recon}, and \Tref{table:ablation-pe} of the manuscript.


\section{Point receptive field analysis}
\label{sec:receptive}
\begin{figure}[!t]
\centering
\includegraphics[width=1.00\linewidth]{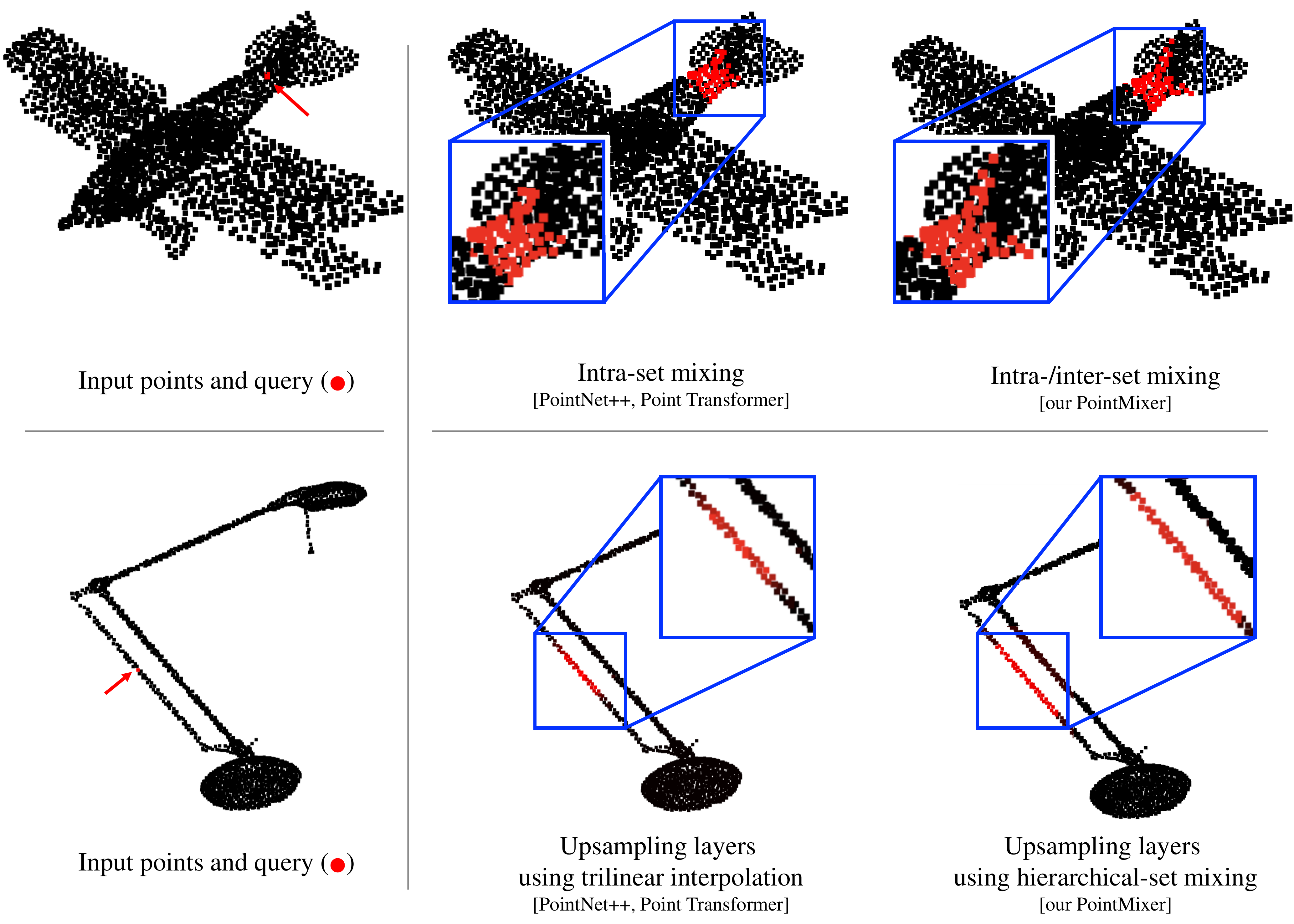}
\caption{Point receptive fields. Given a query point ($\textcolor{red}{\newmoon}$), we colorize the neighbor points that are affected by the query: intra/inter-set (top) and hier-set (bottom).}
\label{fig:fig_attn}
\end{figure}

Throughout this paper, we emphasize the importance of information sharing among unstructured point clouds. As a universal point set operator, PointMixer layer can function as intra-, inter-, and hierarchical-set mixing while previous studies~\cite{pointnet++,point-transformer,xu2021paconv} only focus on the intra-set mixing operations as shown in~\Tref{table:compare} of the manuscript.
For your visual understanding, let us illustrate the receptive fields of our PointMixer layers in different usages as in~\Fref{fig:fig_attn}.
Given a query point~($\textcolor{red}{\newmoon}$), we visualize the receptive fields of intra-set/inter-set mixing (top row) and upsampling layers (bottom row). In particular, we compare trilinear-based upsampling layers (PointNet\plusplus and Point Transformer) and hierarchical-set mixing from our PointMixer layer. We colorize points as red if the red query point influences these. The PointMixer layer has overall larger receptive fields than previous studies~\cite{pointnet++,point-transformer}, which further facilitates point response propagation.

\noindent \textbf{Top row in~\Fref{fig:fig_attn}.} \ 
As stated in~\Sref{subsec:Point Mixer Layer} of the manuscript, intra-set mixing is limited to $k$ closest neighbor points. However, combined use of intra-set and inter-set mixing can propagate point responses into variable length of more neighbors from the neighboring point sets $\mathcal{M}_j$, which is consistently supported by~\Fref{fig:fig_attn}.

\noindent \textbf{Bottom row in~\Fref{fig:fig_attn}.} \ 
Our PointMixer network constitutes a symmetric encoder-decoder network while previous studies do not, as illustrated in~\Fref{fig:fig_symmetric} of the manuscript.
In particular, since creating a new $k$NN graph with $k=16$ is expensive, the trilinear interpolation in upsampling layers of PointNet\plusplus~\cite{pointnet++} and Point Transformer~\cite{point-transformer} usually interpolates three nearest neighbor points, which is the much smaller number of neighbors than 16 in the downsampling layer, and limits their receptive fields as well.
On the other hand, our PointMixer re-uses the $k$NN graph of the downsampling layer to maximize the receptive fields without additional computational costs.
As a result, PointMixer layer can encode point responses in larger contexts than previous approaches~\cite{pointnet++,point-transformer} as shown in \Fref{fig:fig_attn}.
Moreover, these results can be reasons for our superior performance in dense prediction tasks compared to the previous state-of-the-art methods~\cite{point-transformer,point-recon}.


\section{Limited cardinality issues in token-mixing MLPs}
\label{sec:tokenMLPs}
There are two dominant reasons that we remove token-mixing MLPs from our PointMixer layer: limited cardinality and permutation-variant property.
In this section, we further describe the technical reasons of the limited cardinality in token-mixing MLPs.

While a given pixel in an image systematically admits eight adjacent pixels, each point can have an arbitrary number of neighbors in a point cloud. In this context, shared MLPs are particularly desirable since they can handle variable input lengths~\cite{res-mlp}. However, in the vanilla MLP-Mixer layer, token-mixing MLPs limit the process of an arbitrary number of points.

Let us briefly explain the reason. Channel-mixing MLPs require the pre-defined dimensionality in channels for the affine transformation of the input data. 
In contrast, token-mixing MLPs (\Eref{eq:mix_spatial} of the manuscript) switch the channel axis and spatial axis, which results in the pre-defined the number of input tokens (\eg, points). Accordingly, we can only take the pre-defined number of points with fixed channel length as an input of token-mixing MLPs. Thus, token-mixing MLPs cannot operate inter-set mixing whose cardinality varies depending on the point cloud distribution.

\section{Training details}
\label{sec:Training details}
\noindent \textbf{Semantic segmentation} \
We set batch size as 2. Each batch consists of 40K points. We initially set the learning rate as 0.1 and decrease the initial learning rate 10 times smaller at 40, 50 epochs. In total, we train our network for 60 epochs. We use two NVIDIA 1080-Ti GPUs for training. The total training time takes 44 hours. 

\noindent \textbf{Point cloud reconstruction} \
We set batch size as 4. The rest of the training conditions are identical to that of semantic segmentation. To train our network, we modify the header layer of the network that we used for semantic segmentation task. Specifically, we change the channel dimensionality of the output as $3$ that represents $[x,y,z]$.

\noindent \textbf{Object classification} \
We set batch size as 32. We train our network for 300 epochs and schedule the learning rate using cosine-annealing decay. We use the same SGD optimizer that we used in the semantic segmentation task. In our header network, we utilize MLPs with dropout layers and set the ratio as 0.5.

\section{Ablation study (rebuttal)}
In this section, we provide the more ablation studies requested by the reviewers. 

\noindent \textbf{PointMixer vs. previous studies for 3D points.} \
We agree that there are similarities between PointMixer and existing methods (\eg, PointNet\plusplus~\cite{pointnet++}, PointConv~\cite{wu2019pointconv}, and Point Transformer~\cite{point-transformer}) when it comes to the \textit{intra-set} mixing only.
However, in this paper, we aim (1) to improve the network expressiveness via complementary set operations (\textit{intra/inter/hier-set} mixing), 
(2) to develop a universal set operator, and (3) to design a symmetric network using PointMixer.

We integrate \textit{inter/hier-set} mixing blocks into other existing backbones, and compare those variants with our PointMixer as shown in~\Tref{table:diff-backbone}.
Note that PointNet\plusplus and Point Transformer use $\operatorname{max}$ and $\operatorname{vector-attention}$\footnote{Since $\operatorname{vector-attention}$ with an inverse mapping requires many $\operatorname{scatter}$ operations, it also heavily consumes GPU memory.} for \textit{intra-set} mixing, respectively.
The results show that \textit{inter/hier-set} mixing itself consistently improves the performance of PointNet\plusplus and Point Transformer\footnote{Since there are no available pre-trained weights of Point Transformer on both S3DIS and ModelNet40, we trained the Point Transformer ourselves with the official codes provided by the Point Transformer authors.} regardless of the block designs. Interestingly, our mixing scheme~($\operatorname{softmax}$) seems to be more effective than the simple operator ($\operatorname{max}$) as well as the complex  layer~($\operatorname{vector-attention}$).

\noindent\textbf{PointMixer as a 3D version of MLP-Mixer.} \
In~\cite{mlp-mixer},
\textit{“MLP-Mixer contains two types of layers: one with MLPs applied independently to image patches (i.e.,~“mixing” the per-location features), and one with MLPs applied across patches (i.e.,~“mixing” spatial information).”}. Similarly, Synthesizer~\cite{synthesizer} also claims that ``\textit{we show that Random Synthesizers are a form of MLP-Mixers. Random Synthesizers apply a weight matrix on the length dimension as a form of projection across the dimension.}''. 
Based on theses concepts, PointMixer can be seen as a form of both MLP-Mixer and
Synthesizer through $\operatorname{softmax}$ that acts as a projection across the token dimensions, \ie,~token-mixing.

Moreover, MLP-Mixer variants\cite{cyclemlp,vision-permutator,as-mlp,spatial-sift-mlp-v2} also focus on the improved token-mixer.
For example, CycleMLP~\cite{cyclemlp} samples pixels in a cyclic style for linear complexity in token-mixing parts.
AS-MLP~\cite{as-mlp} \textbf{also removes token-mixing MLPs}, and proposes Axial Shift operations for a better local token communication.
From these token-mixing analyses, the direction of PointMixer is aligned with that of MLP-Mixer variants.
Therefore, we respectively argue that PointMixer is a 3D version of MLP-Mixer\footnote{In this context, PointNet\plusplus also can be seen as one of the simplest versions of MLP-Mixer for point cloud understanding. However, as shown in~\Tref{table:diff-backbone}, it is outperformed by PointMixer in the set mixing schemes.}.

\noindent \textbf{Softmax function is not new.} \
Nonetheless, our paper revisits the existing module to emphasize the extended use of $k$NN graph structure.
While previous studies focus on the directional $k$NN graph (\textit{intra-set} mixing), PointMixer newly notices the `bi'-directional characteristics of $k$NN (\textit{inter/hier-set} mixing). Furthermore, to fully utilize this newly-revisited property, the $\operatorname{softmax}$ function can be one choice instead of using complex modules.
In conclusion, our design choice (replacement token-mixing MLPs with $\operatorname{softmax}$ function) is supported by our analysis of permutation-invariant point set operators (\Sref{subsec:MLP-Mixer as Point Set Operator}) across many recent publications~\cite{as-mlp,metaformer,point-transformer} and our extensive experiments (\Tref{table:ablation-pe} and~\Tref{table:diff-backbone}). 
\begin{table}[!t]
\caption{Semantic segmentation results of existing methods with the proposed inter-/hier-set mixing schemes on the S3DIS dataset. Note that `$\operatorname{max}$' represents PointNet\plusplus block using maxpool, `$\operatorname{attn}$' means the vector-attention based Point Transformer block, and `$\operatorname{softmax}$' implies our PointMixer block. All methods are trained for 30 epoch.}
\vspace{-1mm}
\label{table:diff-backbone}
\centering
\resizebox{0.8\linewidth}{!}{
\begin{tabular}{l|ccll}
\toprule

Method & Intra & Inter/Hier & ~~Param. (M)~~ & mIoU $(\%)$ \\

\midrule

\rowcolor{gray!20} PointNet\plusplus & $\operatorname{max}$ & \redxmark & ~~2.0 & 57.3 \\

& $\operatorname{max}$ & $\operatorname{max}$ & ~~2.3 (\textcolor{purple}{$\uparrow$ 0.3}) & 62.7 (\textcolor{teal}{$\uparrow$ 5.4}) \\ 

& $\operatorname{max}$ & $\operatorname{attn}$ & ~~8.3 (\textcolor{purple}{$\uparrow$ 6.3}) & 57.8 (\textcolor{teal}{$\uparrow$ 0.5}) \\

& $\operatorname{max}$ & $\operatorname{softmax}$ & ~~2.7 (\textcolor{purple}{$\uparrow$ 0.7}) & 66.9 (\textcolor{teal}{$\uparrow$ 9.6}) \\

\midrule

\rowcolor{gray!20} PointTransformer$^{\textcolor{red}{10}}$ & $\operatorname{attn}$ & \redxmark & ~~7.8 & 70.0 \\

& $\operatorname{attn}$ & $\operatorname{max}$ & ~~8.1 (\textcolor{purple}{$\uparrow$ 0.3}) & 70.2 (\textcolor{teal}{$\uparrow$ 0.2}) \\

& $\operatorname{attn}$ & $\operatorname{attn}$ & $\!$ 14.1 (\textcolor{purple}{$\uparrow$ 6.3}) & 70.1 (\textcolor{teal}{$\uparrow$ 0.1}) \\

& $\operatorname{attn}$ & $\operatorname{softmax}$ & ~~8.5 (\textcolor{purple}{$\uparrow$ 0.7}) & 70.3 (\textcolor{teal}{$\uparrow$ 0.3}) \\

\midrule

\rowcolor{yellow!20} PointMixer (ours) & ~$\operatorname{softmax}$~ & $\operatorname{softmax}$ & ~~6.5 & 71.4 \\

\bottomrule
\end{tabular}
}
\end{table}


\bibliographystyle{splncs04}
\bibliography{egbib}
\end{document}